\documentclass[5p]{elsarticle}

\pdfoutput=1
\newcommand{\et}{\textit{et al. }}
\usepackage{amsmath,amssymb} % define this before the line numbering.
\usepackage{graphicx}
\usepackage{pdfpages}
\usepackage{url}
\usepackage{xcolor}
\definecolor{newcolor}{rgb}{.8,.349,.1}
\begin{document}

\begin{frontmatter}

\title{Multi-feature Fusion for Image Retrieval Using Constrained Dominant Sets}

%\tnotetext[mytitlenote]{Fully documented templates are available in the elsarticle package on \href{http://www.ctan.org/tex-archive/macros/latex/contrib/elsarticle}{CTAN}.}

%% Group authors per affiliation:
%\author{Elsevier\fnref{myfootnote}}
%\address{Radarweg 29, Amsterdam}
%\fntext[myfootnote]{Since 1880.}

%% or include affiliations in footnotes:

\cortext[mycorrespondingauthor]{Corresponding author}
\author[mymainaddress]{Alemu Leulseged Tesfaye\corref{mycorrespondingauthor}}
\ead{leulseged.alemu@unive.it}

\author[mymainaddress,mysecondaryaddress]{Marcello Pelillo}
%\ead{pelillo@unive.it}

\address[mymainaddress]{DAIS, Ca' foscari University of Venice, Via Torino 155, Mestre, Venezia, Italy}
\address[mysecondaryaddress]{ECLT, European Center for Living Technology, S. Marco 2940, 30124 Venezia, Italy }

\begin{abstract}
Aggregating different image features for image retrieval has recently shown its effectiveness. While highly effective, though, the question of how to uplift the impact of the best features for a specific query image persists as an open computer vision problem. In this paper, we propose a computationally efficient approach to fuse several hand-crafted and deep features, based on the probabilistic distribution of a given membership score of a constrained cluster in an unsupervised manner. First, we introduce an incremental nearest neighbor (NN) selection method, whereby we dynamically select k-NN to the query. We then build several graphs from the obtained NN sets and employ constrained dominant sets (CDS) on each graph G to assign edge weights which consider the intrinsic manifold structure of the graph, and detect false matches to the query. Finally, we elaborate the computation of feature positive-impact weight (PIW) based on the dispersive degree of the characteristics vector. To this end, we exploit the entropy of a cluster membership-score distribution. In addition, the final NN set bypasses a heuristic voting scheme. Experiments on several retrieval benchmark datasets show that our method can improve the state-of-the-art result.
\end{abstract}

\begin{keyword}
Image retrieval\sep multi-feature fusion\sep diffusion process.
\end{keyword}

\end{frontmatter}

%\linenumbers

\section{Introduction}

The goal of semantic image search, or content-based image retrieval (CBIR), is to search for a query image from a given image dataset. This is done by computing image similarities based on low-level image features, such as color, texture, shape and spatial relationship of images. Variation of images in illumination, rotation, and orientation has remained a major challenge for CBIR. Scale-invariant feature transform (SIFT) \cite{Lowe04} based local feature such as Bag of words (BOW) \cite{SivicZICCV03}, \cite{JainBJG12}, \cite{YangNXLZP12}, has served as a backbone for most image retrieval processes. Nonetheless, due to the inefficiency of using only a local feature to describe the content of an image, local-global feature fusion has recently been introduced.

\begin{figure*}[t]

	\includegraphics[width=1\linewidth ,trim=0cm 3.5cm 0cm 0cm,clip]{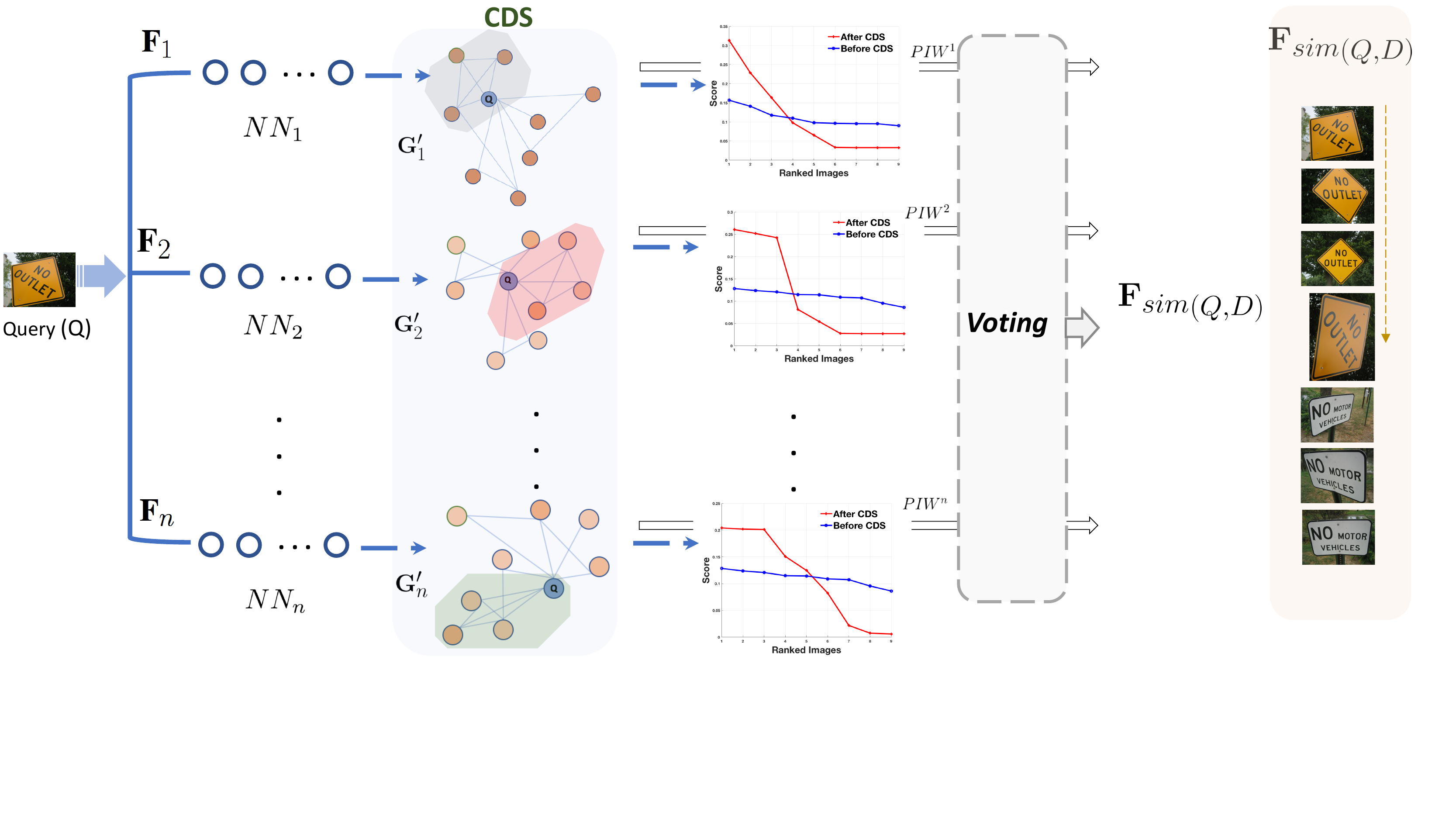}
	
	\caption{ Overview of the proposed image retrieval framework. Based on the given features, $F_1, F_2, . . . F_n,$ we first incrementally collect the $NN's$ to the query $Q,$ denoted as $NN_1, NN_2, . . . NN_n.$ Next, for each $NN$ we build the corresponding graph $G'_1, G'_2, . . . G'_n,$ and then, we apply $CDS$ on each graph to learn the $PIW$ of each feature, $PIW_1, PIW_2, . . . PIW_n,$ in the subsequent plot, the blue and red curves depict the ranked score of NN's before and after the application of CDS,  respectively. Following, the final candidates, which come from each feature, pass through a voting scheme. Finally, using the obtained votes and PIW's we compute the final similarity, $F_{sim} (Q, D)$, between the query and the dataset images by equ. \ref{equ:FinSim} .}
	\label{fig:Workflow}
\end{figure*}

Multi-feature based CBIR attacks the CBIR problem by introducing an approach which utilizes multiple low-level visual features of an image. Intuitively, if the to-be-fused feature works well by itself, it is expected that its aggregation with other features will improve the accuracy of the retrieval. Nevertheless, it is quite hard to learn in advance the effectiveness of the to-be-fused features for a specific query image. Different methods have recently been proposed to tackle this problem \cite{YangMD15}, \cite{ZhaYanCorYuMetPAMI2015}, \cite{ZheWanTiaHeLiuTiaCVPR2015}. Zhang et al. \cite{ZhaYanCorYuMetPAMI2015} developed a graph-based query specific fusion method, whereby local and global rank lists are merged with equal weight by conducting a link analysis on a fused graph.
Zheng et al. \cite{ZheWanTiaHeLiuTiaCVPR2015} proposed a score level fusion model called Query Adaptive Late Fusion (QALF) \cite{ZheWanTiaHeLiuTiaCVPR2015}, in which, by approximating a score curve tail with a  reference collected on irrelevant data, they able to estimate the effectiveness
of a feature as negatively related to the area under the normalized curve. Yang \et\cite{YangMD15} used a mixture Markov model to combine given graphs into one. Unlike \cite{ZhaYanCorYuMetPAMI2015} where graphs are equally weighted, \cite{YangMD15} proposed a method to compute a weight which quantifies the usefulness of the given graph based on a naive Bayesian formulation, which depends only on the statistics of image similarity scores. 

However, existing multi-feature fusion methods have different drawbacks. For instance, \cite{ZheWanTiaHeLiuTiaCVPR2015}, \cite{ZhaYanCorYuMetPAMI2015}, \cite{DenJiLiTaGaICCV2013}, \cite{ZhangYWLT13} heavily rely on a pre-calculated and offline stored data, which turns out to be computationally expensive when new images are constantly added to the dataset. On the other hand, Ensemble Diffusion (ED)\cite{Bai17} requires $O(n^3)$ to perform a similarity diffusion. In addition to that, its feature-weight computation approach is not a query specific. 

Inspired by \cite{ZheWanTiaHeLiuTiaCVPR2015}, in this work we present a novel and simple CBIR method based on a recently introduced constrained cluster notion. Our approach presents two main advantages. Firstly, compared to the state of the art methods, it can robustly quantify the effectiveness of features for a specific query, without any supervision. Secondly, by diffusing the pairwise similarity between the nearest neighbors, our model can easily avoid the inclusion of false positive matches in the final shortlist. Towards this end, we first dynamically collect the nearest neighbors to the query, therefore, for each feature, we will have a different number of NNs. Subsequently, we set up the problem as finding a cluster from the obtained NNs, which is constrained to contain the given query image. To this end, we employ a graph-theoretic method called constrained dominant sets \cite{ZemeneP16}. Here is our assumption: if the nearest neighbor to the query image is a false match, after the application of CDS its membership score to the resulting constrained cluster should be less than the fixed threshold $\zeta,$ which leads us to detect and exclude the outliers. Furthermore, we introduce the application of entropy to quantify the effectiveness of the given features based on the obtained membership score. In contrast to \cite{ZheWanTiaHeLiuTiaCVPR2015}, our method does not need any reference or external information to learn a query specific feature-weight. Fig. \ref{fig:Workflow} shows the pipline of the proposed method.

In particular, we make the following contributions. 1) Compared to the previous work \cite{ZhaYanCorYuMetPAMI2015}, \cite{ZheWanTiaHeLiuTiaCVPR2015}, we propose a simple but efficient entropy-based feature effectiveness weighting system; in addition to that, we demonstrate an effective way of outlier or false nearest neighbor detection method. 2) Most importantly, our proposed model is a generic approach, which can be adapted to distinct computer vision problems, such as object detection and person re-identification. 3) We show that our unsupervised graph fusion model easily alleviates the asymmetry neighborhood problem. 

This paper is structured as follows. In section 2 we briefly survey literature relevant to our problem, followed by technical details of the proposed approach in Sec. 3. And, in Sec. 4 we show the performance of our framework on different benchmark datasets.

%%%%%%%%%%%%%%%%%%%%%%%%%%%

\section{Related Work}

CBIR has become a well-established research topic in the computer vision community. The introduction of SIFT feature plays a vital role in the application of BOW model on the image retrieval problem. Particularly, its robustness in dealing with the variation of images in scale, translation, and rotation provide a significant improvement in the accuracy of similar image search. Sivic et al. \cite{SivicZICCV03} first proposed BOW-based image retrieval method by using SIFT, in that, local features of an image are quantized to visual words. Since then, CBIR has made a remarkable progress by incorporating k-reciprocal neighbor \cite{QinGamBosQuaGooCVPR2011}, query expansion \cite{ChumMPM11}, \cite{QinGamBosQuaGooCVPR2011}, \cite{ToliasJ14}, large visual codebook \cite{PhilbinCISZ07}, \cite{AvrithisK12}, diffusion process \cite{YangMD15} \cite{ZemeneAP16} and spacial verification \cite{PhilbinCISZ07}.  Furthermore, several methods, which consider a compact representation of an image to decrease the memory requirement and boost the search efficiency have been proposed. Jegou et al. \cite{JegouPDSPSPAMI12} developed a Vector of Locally Aggregated Descriptor(VLAD), whereby the residuals belonging to each of the codewords are accumulated.

While SIFT-based local features have considerably improved the result of image search, it does not leverage the discriminative information encoded in the global feature of an image, for instance, the color feature yields a better representation for smooth images. This motivates the introduction of multiple feature fusion for image retrieval. In \cite{ZhaYanCorYuMetPAMI2015}, a graph-based query specific fusion model has been proposed, in which multiple graphs are combined and re-ranked by conducting a link analysis on a fused graph. Following, \cite{YangMD15} developed a re-ranking algorithm by fusing multi-feature information, whereby they apply a locally constrained diffusion process (LCDP) on the localized NNs to obtain a consistent similarity score.

Although the aggregation of handcrafted local and global features has shown promising results, the advent of a seminal work by A.Krizhevsky \et \cite{KrizhevskySH12} in 2012 changed the focus of the computer vision community. Since then, convolutional neural network (CNN) feature has been used as a main holistic cue in different computer vision problems, including CBIR. Despite its significant improvement on the result of image retrieval, CNN feature still can not endow the demanded accuracy on different benchmark retrieval datasets, especially without the use of fine-tuning. Thus, aggregating graphs which are built from a hand-engineered and CNN-based image features has shown improvement in the accuracy of the retrieval \cite{SivicZICCV03}, \cite{JegouPDSPSPAMI12}, \cite{PerronninDCVPR07}, \cite{JegouZCVPR14}, \cite{DoTCCVPR15}, \cite{KalantidisMO16}.

In addition to that, Yang \et \cite{YangMD15} applied a diffusion process to understand the intrinsic manifold structure of the fused graph. Despite a significant improvement on the result, employing the diffusion process on the final (fused) graph restricts the use of the information which is encoded in the pairwise similarity of the individual graph. Instead, our proposed framework applies CDS on each graph which is built from the corresponding feature. Thus, we are able to propagate the pairwise similarity score throughout the graph. Thereby, we exploit the underutilized pairwise similarity information of each feature and alleviate the negative impact of the inherent asymmetry of a neighborhood.

\section{Proposed Method}

\subsection{Incremental NN Selection}

In this subsection, we show an incremental nearest neighbor collection method to the given query image. We start with an intuitive clustering concept that similar nodes with common features should have an approximate score distribution, while outliers, or nodes which do not belong to a similar semantic class, have different score values. Accordingly, we propose a technique to search for the transition point where our algorithm starts including the outlier nodes. To this end, we examine how distinctive two subsequent nodes are in a ranked list of neighbors. Thus, we define a criterion called neighbors proximity coefficient$(NPC),$ which is defined as the ratio of two consecutive NNs in the given ranked list. Therefore, images are added only if the specified criterion is met, which is designed in such a way that only images that are very likely to be similar to the query image are added. Thereby, we are able to decrease the number of false matches to the query in the k-nearest neighbors set. 

Given an initial ranked list $R$. And then, we define top-k nearest neighbors (kNN) to query $Q$ as

\begin{equation}
	kNN(q, k) = 
	\begin{cases}
		\textrm{Add} \: n_i & if  \quad   \frac{Sim(q,n_{i+1})}{Sim(q,n_{i})} > $NPC$    \\
		0 & otherwize  
	\end{cases}
\end{equation} 

where $ |kNN(q, k)| = k,$ and $|.|$ represents the cardinality of a set.

\begin{equation}
kNN(q, k)= \{n_1, n_2, . . . n_k\},\quad where \; kNN(q,k) \subseteq R
\end{equation}

\subsection{Graph Construction}\label{PIWusingCDS}

Different features, $F = F_1, F_2 . . . F_n,$ are extracted from images in the dataset D and the query image $Q$, where each feature encodes discriminative information of the given image in different aspects. We then compute the distance between the given images based on a distance metric function $d'(I_i,I_j),$ where $I_i$ and $I_j$ denote the given feature vector extracted from image $i$ and $j$ respectively. Following, we compute symmetric affinity matrices $A_1'$, $A_2'$, . . . $A_n'$  from each distance matrix $D_i$ using a similarity function $S(D_i)$. We then apply minimax normalization on each similarity matrix as: $	A_i= \frac{V_\alpha^{ij} - min(V_\alpha)}{max(V_\alpha) - min(V_\alpha)}$, where $V_\alpha$ is a column vector taken from matrix $A'_i$, which comprises the pairwise similarity score between a given image $V_\alpha^i$ and images in the dataset $V^j,$ which is denoted as $V_\alpha^{ij}$. Next, we build undirected edge-weighted graphs with no self-loops $G_1, G_2 . . . G_n$ from the affinity matrices $A_1, A_2, . . . A_n,$ respectively. Each graph $G_n$  is defined as $Gn = (V_n, E_n, w_n),$ where $V_n = { 1, . . . ,n }$ is vertex set, $E_n \subseteq V_n \times V_n$ is the edge set, and $w_n: E \longrightarrow {\rm I\!R}_+^*$  is the (positive) weight function. Vertices in G correspond to the given images, edges represent neighborhood relationships, and edge-weights reflect similarity between pairs of linked vertices.

\subsection{PIW Using Entropy of CDS}

Since the nearest neighbor selection method heavily relies on the initial pairwise similarity, it is possible that the NN set easily includes false matches to the given query. This usually happens due to the lack of technics which consider the underlying structure of the data manifold,  especially the inherent asymmetry of a neighborhood is a major shortcoming of such systems. For instance, although $Sim(n_i, q) = Sim(q, n_i),$ the nearest neighbor relationship between query $Q$ and image $n_i$ may not be symmetric, which implies that  $m_i \in kNN(q, k)$ but $m_i \notin kNN(n_i, k).$
As demonstrated in the past retrieval works, the k-reciprocal neighbors \cite{QinGamBosQuaGooCVPR2011} and similarity diffusion process \cite{IscenTAFC17} have been vastly taken as the optimal options to tackle this issue. However, the existing methods are not computationally efficient. In this work, we remedy the existing limitations using an unsupervised constrained clustering algorithm whereby we exploit the pairwise similarity to find a cohesive cluster which incorporates the specified query.

%\subsubsection*{Constructing $G'$ from RN}

\subsubsection{Constrained Clustering for Coherent Neighbor Selection}\label{kRN}

\begin{figure*}[t]
	
	\centering
	%content...%\fbox{\rule{0pt}{2in} \rule{0.9\linewidth}{0pt}}
	
	\includegraphics[width=1\linewidth ,trim=0cm 6.8cm 0cm 0cm,clip]{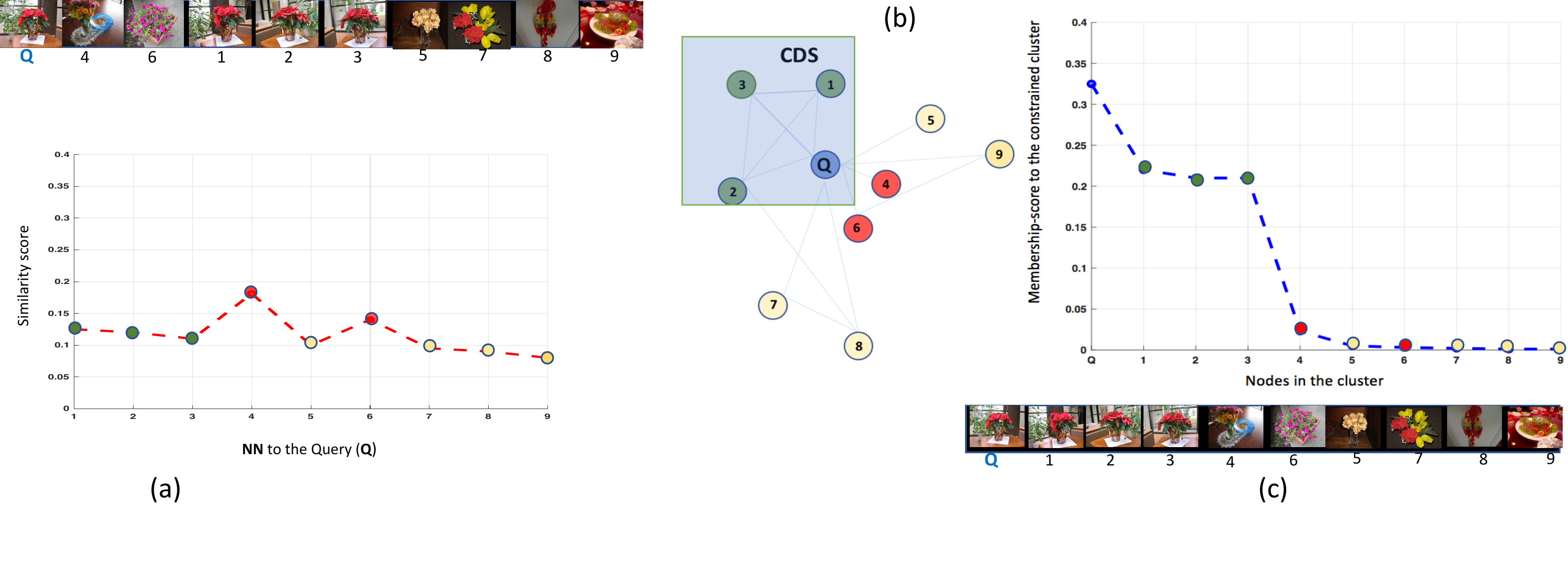}
	
	\caption{(a) Initial score distribution of the top k nearest neighbors to the query Q, green and red points denote the false-negative and false-posetive NNs. (b) Graph $G'$, built from the initial pairwise similarity of the k-nearest neighbor set. And the blue box contains the CDS nodes which are obtained by running CDS on graph $G'.$ (c) The resulting constrained dominant set membership-score distribution.}
	\label{fig:CharacvecExam}
\end{figure*}

Towards collecting true matches to the query image, we employ an unsupervised clustering algorithm on the top of the previous steps. Our hypothesis to tackle the asymmetry problem between the given query and its nearest neighbors is that images which are semantically similar to each other tend to be clustered in some feature space. As can be seen in the synthetic example (See Fig. \ref{fig:CharacvecExam}), retrieved image $i_4$ and $i_6$ are outliers or false positives to the query image $Q$. We can confirm this by observing the common neighbors of $Q$ with $i_4$ and $i_6$. But due to the lack of contextual information, the system considers them as a true match (neighbor) to the query. In our proposed model, to attack this issue, we represent the set of $kNN$ as a graph $G'$ accordingly to subsection \ref{PIWusingCDS}. Then, we treat outliers finding problem as an unsupervised clustering problem. We first convert graph $G'$ into a symmetric affinity matrix $A$, where the diagonal corresponding to each node is set to 0, and the $ij-th$ entry denotes the edge-weight $w_{ij}$ of the graph so that $A_{ij}\equiv A_{ji}$. Accordingly, given graph $G'$ and query $Q$, we cast detecting outliers from a given $NN$ set as finding the most compact and coherent cluster from graph $G',$ which is constrained to contain the query image $Q.$ To this end, we adopt constrained dominant sets \cite{ZemeneP16}, \cite{ZemeneAP17}, which is a generalization of a well known graph-theoretic notion of a cluster. We are given a symmetric affinity matrix $A$ and parameter  $\mu > 0,$  and then we define the following parametrized quadratic program

\begin{equation} \label{eqn:parQP}
\begin{array}{ll}
\text{maximize }  &  f_Q^\mu(X) = X' (A - \mu \hat \Gamma_Q) X \\

&  f_Q^\mu(X) = X' \hat{A} X \\

\text{subject to} &  X \in \Delta
\end{array}
\end{equation}

where a prime denotes transposition and  
$$
\Delta=\left\{ X \in R^n~:~ \sum_{i=1}^n X_i = 1, \text{ and } X_i \geq 0 \text{ for all } i=1 \ldots n \right\}
$$ 
$\Delta$ is the standard simplex of $R^n$. $\hat \Gamma_Q$ represents $n \times n$ diagonal matrix whose diagonal elements are set to zero in correspondence to the query $Q$ and to 1 otherwise. And $\hat{A}$ is defined as,

$$
\hat{A} = A - \mu \hat \Gamma_Q  = 
\begin{pmatrix} 
0 & .    & .  &.  \\ 
. & ~~ - \mu ~~ &  .  &.\\
%~~ ~~ & ~~ ~~ & ~~ ~~ &   \\
.& .    & - \mu  &. \\ 

%~~ ~~ & ~~ ~~ & ~~ ~~\\
.& .   &  .   & -\mu\\ 
\end{pmatrix}
$$ 
where the dots denote the $ij$ th entry of matrix $A.$ Note that matrix $\hat{A}$ is scaled properly to avoid negative values.

Let $Q \subseteq V,$ with $Q \neq \emptyset$ and let $\mu > \lambda_{max}(A_{V \backslash Q} ),$ where $\lambda_{max}(A_{V\backslash q})$ is the largest eigenvalue of the principal submatrix of $A$ indexed by the element of $V\backslash q.$ If $X$ is a local maximizer of $ f_Q^\mu(X)$ in $\Delta,$ then $\delta(X) \cap Q \neq \emptyset,$ where, $\delta(X) = {i \in V : X_i > 0.}$ We refer the reader to \cite{ZemeneP16} for the proof.

The above result provides us with a simple technique to determine a constrained dominant set which contains the query vertex $Q.$ Indeed, if $Q$ is the vertex corresponding the query image, by setting 

\begin{equation}
\mu > \lambda (A_{V\backslash Q})
\end{equation}
we are guaranted that all local solutions of eq ($\ref{eqn:parQP}$) will have a support that necessarily contains the query element.
The established correspondence between dominant set (coherent cluster) and local extrema of a quadratic form over the standard simplex allow us to find a dominant set using straightforward continuous optimization techniques known as replicator dynamics, a class of dynamical systems arising in evolutionary game theory [21]. The obtained solution provides a principled measure of a cluster cohesiveness as well as a measure of vertex participation. Hence, we show that by fixing an appropriate threshold $\zeta$ on the membership score of vertices, to extract the coherent cluster, we could easily be able to detect the outlier nodes from the k-nearest neighbors set. For each $X^i,$ $\zeta^i$ is dynamically computed as
\begin{equation}
\zeta^i = \Lambda(1 - max(X^i) + min(X^i))/L
\end{equation}

where $max(X)$ and $min(X)$  denote the maximum and minimum membership score of $X^i,$ respectively. $\Lambda$ is a scaling parameter and $L$ stands for length of $X^i.$ Moreover, we show an effective technique to quantify the usefulness of the given features based on the dispersive degree of the obtained characteristics vector $X.$ 

\subsubsection{PIW Using Entropy of Constrained Cluster.}

Entropy has been successfully utilized in a variety of computer vision applications, including object detection \cite{SznitmanBFF13}, image retrieval \cite{DeselaersWN06} and visual tracking \cite{MaLFZ15}. In this paper, we exploit the entropy of a membership-score of nodes in the constrained dominant set to quantify the usefulness of the given features. To this end, we borrowed the concept of entropy in the sense of information theory (Shannon entropy). We claim that the discriminative power of a given feature is inversely proportional to the entropy of the score distribution, where the score distribution is a stochastic vector. Let us say we are given a random variable $C$ with possible values $c_1, c_2, . . . c_n,$ according to statistical point of view the information of the event ($C = c_i$) is inversely proportional to its likelihood, which is denoted by $I(C_i)$ and defined as 

\begin{equation}
I(C_i) = log \Big( \frac{1}{P(c_i)}\Big) = -log (p(c_i)).
\end{equation}
 Thus, as stated by \cite{Shannon}, the entropy of $C$ is the expected value of I, which is given as 
 \begin{equation}
 	H(C) = - \sum_{i=1}^N P(c_i)log(P(c_i).
 \end{equation}
For each characteristic vector {\small $X^i,X^{i + 1} . . . X^z,$} where {\small $X^i = \big\{X^i_\mu, X^i_{\mu + 1}. . . X^i_{n}\big\},$} we compute the entropy $H(exp({X^i}))$. Each $X^i$ corresponds to the membership score of nodes in the CDS, which is obtained from the given feature $F^i$.  Assume that the top NNs obtained from feature x are irrelevant to the query Q, thus the resulting CDS will only contain the constraint element Q. Based on our previous claim, since the entropy of a singleton set is 0, we can infer that the feature is highly discriminative. Although this conclusion is right, assigning a large weight to feature with irrelevant NNs will have a negative impact on the final similarity. To avoid such unintended impact, we consider the extreme case where the entropy is 0. Following, we introduce a new term $C_a,$ which is obtained from the cardinality of a given cluster, $K_c,$ as 	$Ca^i = \frac{K_c ^i}{\sum_{i=1}^z  K_c^i}.$ As a result, we formulate the PIW computation from the additive inverse of the entropy $\varepsilon ^i =  1-H(X^i),$ and $C_a^i,$ as: 
\begin{equation}\label{eq.PIW}
PIW^i = \frac{\vartheta ^i}{\sum_{i=1}^z  \vartheta^i} \quad Thus, \sum_{i=1}^{z} PIW^i = 1
\end{equation}
where $\vartheta ^i = \varepsilon^i + C_a^i,$ and $i$ represents the corresponding feature.

%\begin{equation}
%	sim(q,d)=\prod_{i=1}^k(s_{d,q}^{(i)})^{\omega_q^{(i)}}\quad where\! \sum_{i=1}^k\omega_q^{(i)}=1
%\end{equation}
\subsection{Naive Voting Scheme and Similarity Fusion}

In this section, we introduce a simple yet effective voting scheme, which is based on the member nodes of k-nearest neighbor sets and the constrained dominant sets, let {\small $NN_1, NN_2 . . . NN_z$} and {\small $CDS_1, CDS_2 . . . CDS_z$} represent the $NN$ and $CDS$ sets respectively, which are obtained from $G'_1,G'_2 . . . G'_z.$ Let us say $\xi = 2(z-1) - 1,$ and then  we build $\xi$  different combinations of $NN$ sets, $\varphi_1, \varphi_2 . . . \varphi_{\xi}.$ Each $\varphi$ represents an intersection between $z - 1$ unique combinations of $NN$ sets. We then form a super-set $\varpi$ which contains the union of $\varphi$ sets, with including repeated nodes. Assume that {\small $NNs = \{NN_1, NN_2, NN_3\}, \xi = 3,$} thus each $\varphi$ set contains the intersection of two $NN$ sets as {\small $\varphi_1 = \{NN_1 \cap NN_2\}, \varphi_2 = \{NN_1 \cap NN_3\}$} and {\small $\varphi_3=\{NN_2 \cap NN_3\}.$} Hence the resulting $\varpi$ is defined as $\varpi = (\varphi_1 \ominus \varphi_2 \ominus \varphi_3),$ where $(.\ominus.)$ is an operator which returns the union of given sets, including repeated nodes. We have also collected the union of $CDS$ sets as {\small $\omega = (CDS_1 \ominus CDS_2 \ominus CDS_3).$} Following, we compute $\kappa$ as $(\kappa = \varphi_1 \cap \varphi_2 \cap . . . \varphi_{\xi}).$ Thereby we find super-sets $\varpi, \omega$ and $\kappa.$ Next, we design three different counters, which are formulated to increment when the NN node appears in the corresponding super-sets. Based on the value obtained from each counter, we finally compute the vote scores for each $NN$ node to the query as $v_1 = v_1/\eta, v_2 = v_2/\theta$ and $v_3 = v_3/\iota,$ where $\eta, \theta$ and $\iota$ are parameters which are fixed empirically. Note that the outlier detecting capability of our framework is encoded in the voting process. Thus, if a NN node $n_{i}$ is contained in more than one cluster, its probability to be given a large weight is higher. This is due to the number of votes it gets from each cluster.

\subsubsection{Final Similarity.}

After obtaining the aforementioned terms, we compute the final similarity as follows: say we are given $n$ features, $Q$ is the query image and $D$ denotes image  dataset, then the initial similarity of $D$ to $Q$, with respect to feature $F_i, i = 1 . . . n,$ ,is given as $S_{D, Q}^{(i)}.$ Let $PIW_{Q}^{(i)}, i = 1 . . . n,$ encode the weight of feature $F_{i}$ for query $Q,$ and then the final similarity score, $F _{sim (Q, D)} $, between $Q$ and $D$ is given as

\begin{equation}
N_{s} = \prod_{i = 1}^{k} (S_{D, Q}^{(i)})^{PIW_Q^{(i)}} \\
\end{equation}

\begin{equation}\label{equ:FinSim}
F _{sim (Q, D)} = \lambda N_s + (1 - \lambda) \sum_{\Omega = 1}^{\Psi} v_\Omega
\end{equation}
where $\Psi = 3,$ is the total number of voter sets. And $\lambda \in [0, 1]$ defines the penalty factor which penalizes the similarity fusion, when $\lambda = 1$ only $F_s$ is considered, otherwise, if $\lambda = 0,$ only $v$ is considered. 

%-------------------------------------------------------------------------

\section{Experiments}

In this section, we present the details about the features, datasets and evaluation methodology we used along with rigorous experimental analysis.

\subsection{Datasets and Metrics} 

To provide a thorough evaluation and comparison, we evaluate our approach on INRIA Holiday, Ukbench, Oxford5k and Paris6k datasets. 

\textbf{Ukbench Dataset \cite{NisterS06}.} Contains 10,200 images which are categorized into 2,550 groups, each group consists of three similar images to the query which undergo severe illumination and pose variations. Every image in this dataset is used as a query image in turn while the remaining images are considered as dataset images, in ``leave-one-out" fashion. As customary, we used the N-S score to evaluate the performance of our method, which is based on the average recall of the top 4 ranked images.

\textbf{INRIA Holiday Dataset \cite{JegDouSchECCV2008}.} Comprises 1491 personal holiday pictures including 500 query images, where most of the queries have one or two relevant images. Mean average precision (MAP) is used as a performance evaluation metric.

\textbf{Oxford5k Dataset \cite{PhilbinCISZ07}.} It is one of the most popular retrieval datasets, which contains 5062 images, collected from flicker-images by searching for landmark buildings in the Oxford campus. 55 queries corresponding to 11 buildings are used.

\textbf{Paris6k Dataset \cite{PhilbinCISZ08}.} Consists of 6392 images of Paris landmark buildings with 55 query images that are manually annotated.

\begin{figure*}
	
	\begin{center}
		%content...%\fbox{\rule{0pt}{2in} \rule{0.9\linewidth}{0pt}}
		
		\includegraphics[width=1\linewidth ,trim=0cm 5.5cm 0cm 0cm,clip]{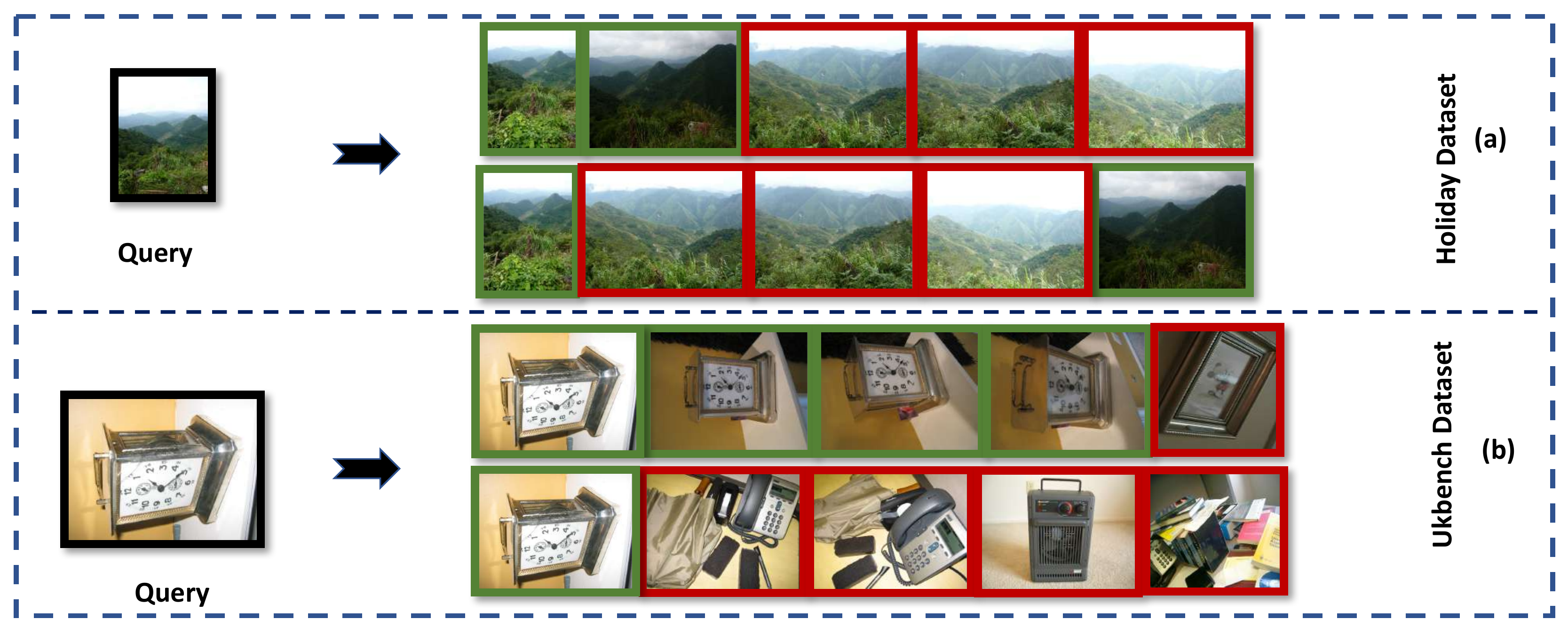}
	\end{center}
	
	\caption{ five relevant images to the query where the green and red frame indicate the True and False posetives to the query, respectively. \textbf{Top-row (a) and (b):} show the top five relevant images of our proposed method. \textbf{Bottom row (a) and (b):} show the top five relevant images obtained from a Naive fusion of several features.}
	\label{fig:HOLex}
\end{figure*}

\subsection{Image Features}

\textbf{Object Level Deep Feature Pooling (OLDFP)\cite{MopuriB15}.} OLDFP is a compact image representation, whereby images are represented as a vector of pooled CNN features describing the underlying objects. Principal Component Analysis (PCA) has been employed to reduce the dimensionality of the compact  representation. We consider the top 512-dimensional vector in the case of the Holiday dataset while considering the top 1024-dimensional vector to describe images in the Ukbench dataset.  As suggested in \cite{MopuriB15}, we have applied power normalization (with exponent 0.5), and l2 normalization on the PCA projected image descriptor.

\textbf{BOW.} Following \cite{ZhengWLT14}, \cite{ZheWanTiaHeLiuTiaCVPR2015}, we adopt Hamming Embedding \cite{JegDouSchECCV2008}. SIFT descriptor and Hessian-Affine detector are used in feature extraction, and we used 128-bit vector binary signatures of SIFT. The Hamming threshold and weighting parameters are set to 30 and 16 respectively, and three visual words are provided for each key-point. Flickr60k data \cite{JegDouSchECCV2008} is used to train a codebook of size 20k. We also adopt root sift as in \cite{ArandjelovicZ12}, average IDF as defined in \cite{ZhengWLT13} and the burstiness weighting \cite{JegouDS09}.

\textbf{NetVLAD \cite{ArandjelovicGTP16}.}  NetVLAD is an end-to-end trainable CNN architecture that incorporates the generalized VLAD layer.

%\textbf{NetVLAD:} is an end-to-end trainable CNN architecture that incorporate the generalized VLAD layer.

\textbf{HSV Color Histogram.} Like \cite{YangMD15}, \cite{ZheWanTiaHeLiuTiaCVPR2015}, for each image, we extract 1000-dimensional HSV color histograms where the number of bins for H, S, V are 20, 10, 5 respectively.

% check the number of bines we used for HSV, to be sure

\begin{table*}[h]
	\centering
	%\begin{center}
	\caption{\small The performance of baseline features on Holidays, Ukbench, Oxford5k and Paris6k datasets. }
	\smallskip
	\begin{tabular}{|c|c|c|c|c|c|c|c|c|c|} 
		%\hline\noalign{\smallskip}
		\hline
		 Datasets & Metrics&  NetVLAD \cite{ArandjelovicGTP16} &  BOW &  OLDFP& HSV& $R_{res}$\cite{IscenTAFC17}&  $G_{res}$\cite{IscenTAFC17}& $R_{vgg}$ \cite{IscenTAFC17} & $G_{vgg}$\cite{IscenTAFC17} \\\hline\hline
		\textbf{Holidays} & MAP& 84&80 &87 &65& -& -&-&-\\
		\textbf{Ukbench} &N-S score& 3.75& 3.58& 3.79&3.19& &-&-&-\\
		\textbf{Oxford5k} &MAP& 69& -&- &-& 95.8&87.7&93&-\\
		\textbf{Paris6k} &MAP& -& -&- &-& 96.8 & 94.1 & 96.4 &95.6\\
		
		\hline
	\end{tabular}
	%\end{center}
	\label{table:ImagePair}
\end{table*}

%\vspace{-1em
%\vspace{8mm}

\subsection{Experiment on Holiday and Ukbench Datasets}

As it can be seen in Fig.\ref{fig:HOLex}(a), the noticeable similarity between the query image and the irrelevant images, in the Holiday dataset, makes the retrieval process challenging. For instance, (See Fig.\ref{fig:HOLex}(a)), at a glance all images seem similar to the query image while the relevant are only the first two ranked images. Moreover, we can observe that the proposed scheme is invariant to image illumination and rotation change. Table \ref{table:comparison} shows that our method significantly improves the MAP of the baseline method \cite{MopuriB15} on Holiday dataset by 7.3 $\%$ while improving the state-of-the-art method by 1.1 $\%$. Likewise, it can be seen that our method considerably improves the N-S score of the baseline method \cite{MopuriB15} on the Ukbench dataset by 0.15 while improving the state-of-the-art method by 0.03. 
\begin{figure*}
	\centering
	%content...%\fbox{\rule{0pt}{2in} \rule{0.9\linewidth}{0pt}}
	\includegraphics[width=1\linewidth ,trim=0cm 0cm 0cm 0cm,clip]{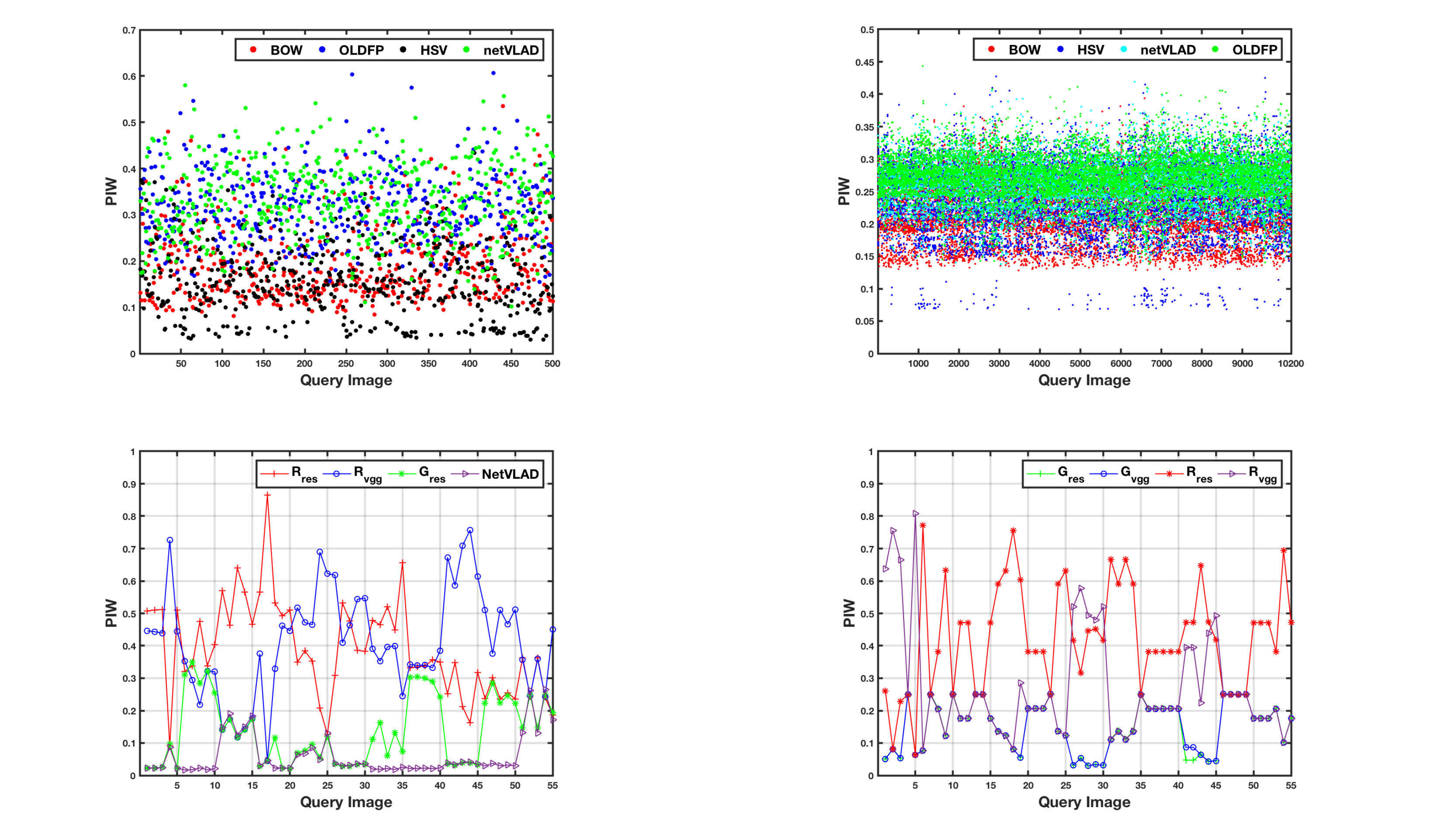}
	\caption{Feature positive-impact weights (PIW's) learned by our algorithm. Top-left, top-right, bottom-left, and bottom-right: on Holiday, Ukbench, Oxford5k and Paris6k datasets, respectively.}
	\label{fig:learnedPIW}
\end{figure*}
%\vspace{8mm}

\begin{figure*}
	\centering
	%content...%\fbox{\rule{0pt}{2in} \rule{0.9\linewidth}{0pt}}
	\includegraphics[width=1\linewidth ,trim=0cm 9.5cm 0cm 0cm,clip]{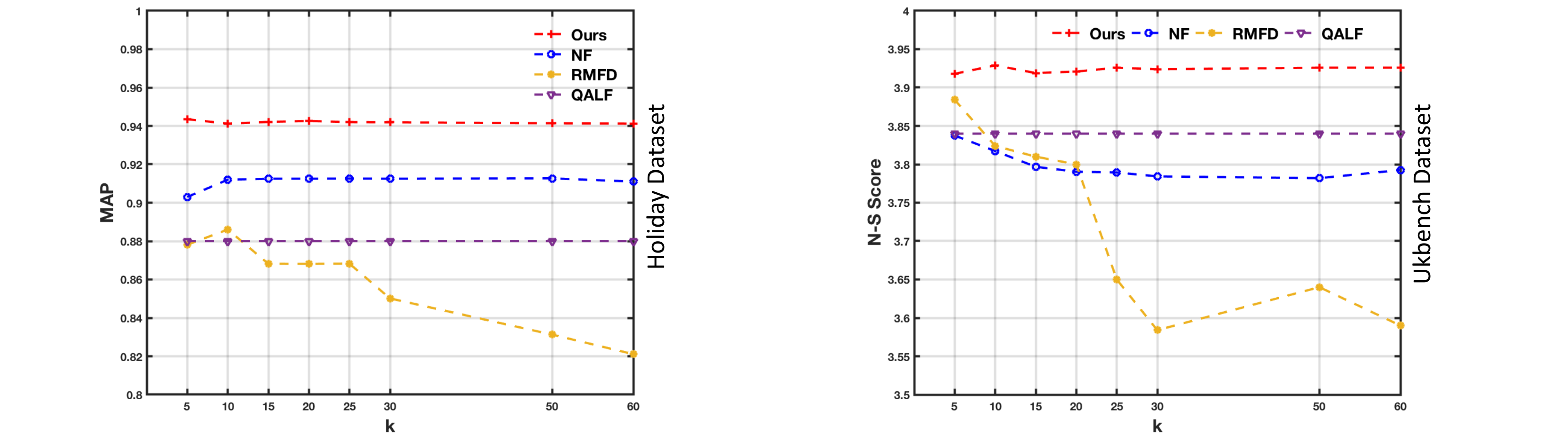}
	\caption{Comparison with state-of-the-art fusion methods with respect to varying k. Naive Fusion (NF), Reranking by Multi-feature Fusion (RMFD) \cite{YangMD15}, and QALF \cite{ZheWanTiaHeLiuTiaCVPR2015}.}
	\label{fig:WithPIWewithourPIW}
\end{figure*}

Furthermore, to show how effective the proposed feature-weighting system is, we have experimented by fusing the given features with and without PIW. Naive fusion (NF) denotes our approach with a constant PIW for all features used, thus the final similarity $F_s$ defined as $F_s = \frac{1}{k} (\prod_{i = 1}^k (S_{D, Q}^{(i)})).$ In Fig.\ref{fig:WithPIWewithourPIW} we have demonstrated the remarkable impact of the proposed PIW. As can be observed, our scheme effectively uplifts the impact of a discriminative feature while downgrading the inferior one. Note that in the PIW computation we have normalized the minimum entropy (See eq.\ref{eq.PIW}), thus its values range between 0 and 1. Accordingly, one implies that the feature is highly discriminative, while zero shows that the feature is indiscriminate.

\begin{figure*}
	\centering
	%content...%\fbox{\rule{0pt}{2in} \rule{0.9\linewidth}{0pt}}
	\includegraphics[width=1\linewidth ,trim=0cm 0cm 0cm 0cm,clip]{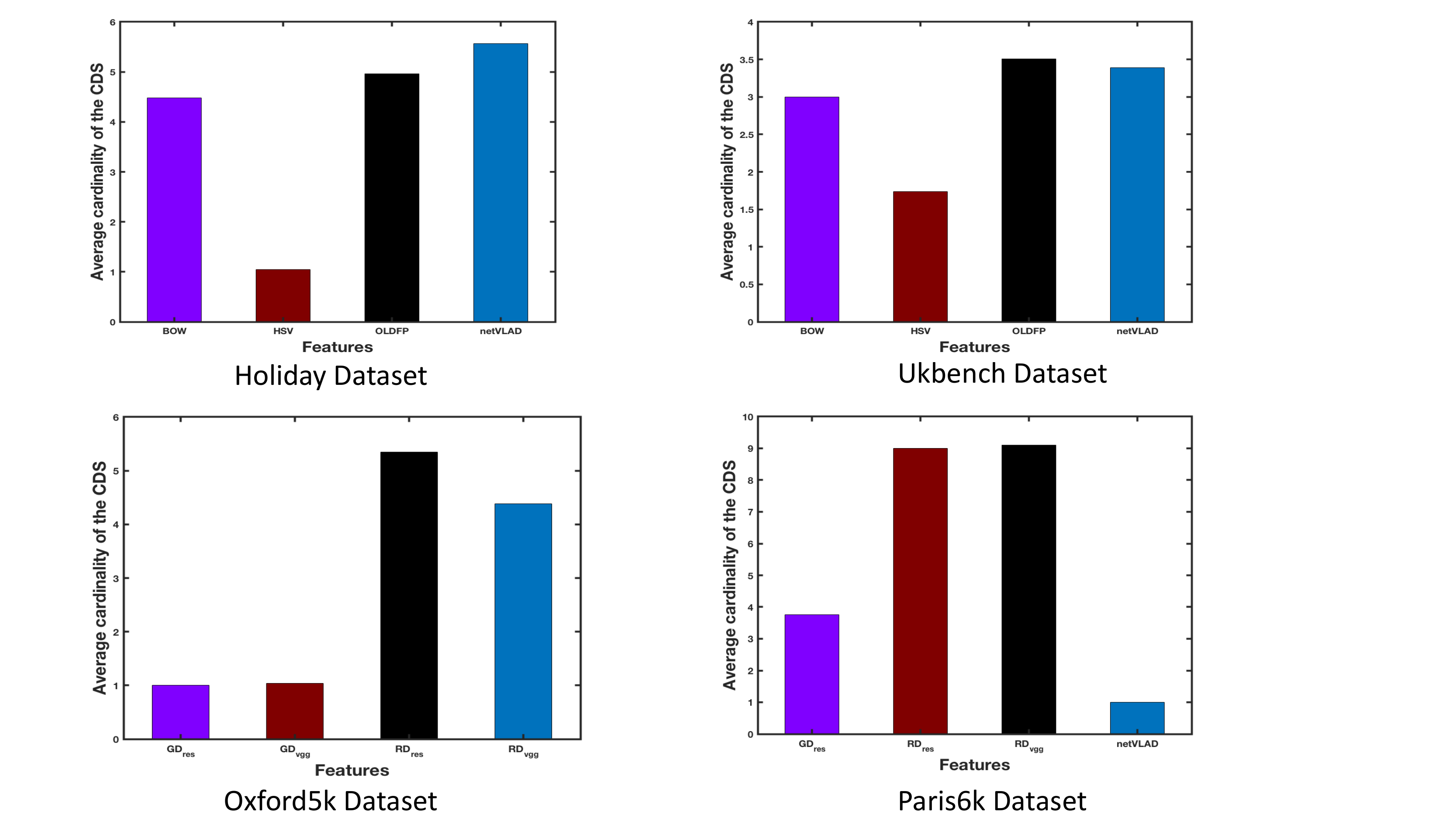}
	%\vspace{5mm}
	\caption{The cardinality of constrained dominant sets for the given features.}
	\label{fig:ClusterCardinality}
\end{figure*}
%\vspace{8mm}
In order to demonstrate that our scheme is robust in handling outliers, we have conducted an experiment by fixing the number of NNs (disabling the incrimental NNs selection) to different numbers. As is evident from Fig.\ref{fig:WithPIWewithourPIW}, the performance of our method is consistent regardless of the number of $kNN$. As elaborated in subsection \ref{kRN}, the robustness of our method to the number of $k$ comes from the proposed outlier detection method. Since the proposed outliers detector is formulated in a way that allows us to handle the outliers, we are easily able to alleviate the false matches which are incorporated in the nearest neighbors set. This results in finding a nearly constant number of nearest neighbors regardless of the choice of $k$.

\begin{table*}[h]
	\centering
	%\begin{center}
	\caption{\small Comparison among various retrieval methods with our method on benchmark datasets, where QALF is implemented with the same baseline similarities used in our experiments.}
	\begin{tabular}{|c|c|c|c|c|c|c|c|c|c|c|c|} 
		%\hline\noalign{\smallskip}
		\hline
		%\hline\noalign{\smallskip}
		Datasets & Metrics& Baselines  & QALF\cite{ZheWanTiaHeLiuTiaCVPR2015} &  \cite{YangMD15} &NF& ED\cite{BaiZWBLT17} & \cite{GordoARL16}&\cite{RadenovicTC16} &\cite{XuSQWX17}&\cite{BabenkoSCL14}&   Ours \\
	
		\hline\hline
		%\noalign{\smallskip}
		\textbf{Ukbench} &\scriptsize{N-S score}& 3.79\cite{MopuriB15} & 3.84 & 3.86 & 3.86&3.93&-&-&-&3.76&\textbf{3.94}      \\
		\textbf{Holiday}   &\scriptsize{MAP}& 87\cite{MopuriB15} &  88& 88&91&93 &90&83&89&77&\textbf{94}       \\   		 
		\textbf{Oxford5k}  &\scriptsize{MAP}&  95.8\cite{IscenTAFC17}&  \textbf{-} & 76.2& 94.4&-&89.1&79.7&81.4&67.6&\textbf{96.2}       \\ 
		\textbf{Paris6k}  & \scriptsize{MAP}& 96.8\cite{IscenTAFC17}&  - & 83.3 & - &-&91.2&83.8&88.9&-& \textbf{97.4}\\ 
		\hline
	\end{tabular}
	%\end{center}
	\label{table:comparison}
\end{table*}

\subsection{Experiment on Oxford5k and Paris6k Datasets}
In the same fashion as the previous analysis, we have conducted extensive experiments on the widely used Oxford5k and Paris6k datasets. Unlike the Holiday and Ukbench datasets, we adapt affinity matrices which are obtained through a diffusion process on a regional $Resnet$ and $VGG$ representation \cite{IscenTAFC17}, and they are denoted as $R_{res}$ and $R_{vgg}$ respectively, as well as affinity matrices $G_{res}$ and $G_{vgg}$ which are also obtained through a diffusion process on a global $Resnet$ and $VGG$ representation, respectively. Table \ref{table:comparison} shows that the proposed method slightly improves the state-of-the-art result. Even if the performance gain is not significant, our scheme marginally achieves better MAP over the state-of-the-art methods. Furthermore, as shown in Fig \ref{fig:learnedPIW}, the proposed model learns the PIW of the given features effectively. Therefore, a smaller average weight is assigned to $G_{vgg}$ and $NetVLAD$ feature comparing to $R_{res}$ and $R_{vgg}$.

\subsection{Robustness of Proposed PIW}
As can be seen in Fig \ref{fig:learnedPIW}, for all datasets, our algorithm has efficiently learned the appropriate weights to the corresponding features. Fig. \ref{fig:learnedPIW} shows how our algorithm assigns PIW in a query adaptive manner. In Holiday and Ukbench datasets, the average weight given to HSV feature is much smaller than all the other features used. Conversely, a large PIW is assigned to OLDFP and NetVLAD features. Nevertheless, it is evident that in some cases a large value of PIW is assigned to HSV and BOW features as well, which is appreciated considering its effectiveness on discriminating good and bad features in a query adaptive manner.

\subsection{Impact of Parameters}

\begin{figure*}
	\centering
	%content...%\fbox{\rule{0pt}{2in} \rule{0.9\linewidth}{0pt}}
	\includegraphics[width=1\linewidth ,trim=0cm 0cm 0cm 0cm,clip]{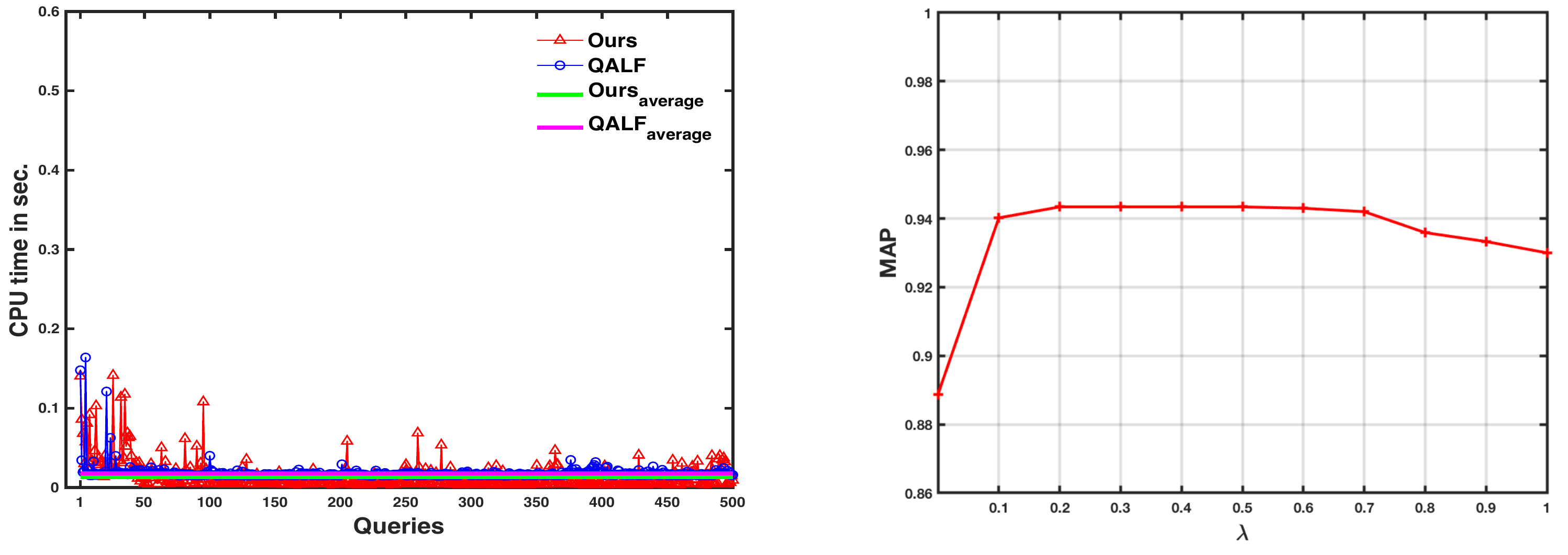}
	%\vspace{5mm}
	\caption{Left: Time complexity of our algorithm (red) and QALF\cite{ZheWanTiaHeLiuTiaCVPR2015} (blue) on Holiday dataset. Right: The impact of $\lambda$ on the retrieval performance, on Holiday dataset.}
	\label{fig:ComplexityeLambda}
\end{figure*}

To evaluate the robustness of our method we have performed different experiments by changing one parameter at a time. Thereby, we have observed that setting  $\Lambda$ to a large value results in assigning insignificant PIW to indiscriminate features. The reason is that after the application of CDS, the cluster membership-score of the dissimilar images to the query will become smaller. Thus, since the threshold fixed to choose the true neighbors is tighter, the resulting constrained dominant set will be forced  to yield a singleton cluster. As a result, we obtained a very small PIW due to the cardinality of the constrained-cluster. In addition to that, we observe that the MAP start to decline when $\lambda$ is set to a very large value (See. Fig \ref{fig:ComplexityeLambda}, right).

\subsection{Impact of Cluster Cardinality}
On the Ukbench dataset, as can be observed in Fig. \ref{fig:ClusterCardinality}, the average cardinality of the constrained clusters which is obtained from HSV and BOW feature is 3 and 1.7, respectively. In contrast, for NetVLAD and OLDFP, the average cluster cardinality is 3.4 and 3.5, respectively . Similarly, in the case of the Holiday dataset, the cluster cardinality obtained from HSV feature is one while for BOW, NetVLAD and OLDFP is 4.5, 5 and 5.6, respectively. Thus, from this, we can draw our conclusion that the cardinality of a constrained dominant set, in a certain condition, has a direct relationship with the effectiveness of the given feature.

\subsection{Computational Time}
In Fig. \ref{fig:ComplexityeLambda} we depict the query time taken to search for each query image, red and blue lines represent our method and QALF, respectively. The vertical axis denotes the CPU time taken in seconds, and the horizontal axis shows the query images. As can be seen from the plot, the proposed framework is faster than the fastest state-of-the-art feature-fusion method \cite{ZheWanTiaHeLiuTiaCVPR2015}. As for time complexity, in our experiment we used a replicator dynamics to solve problem (\ref{eqn:parQP}), hence, for a graph with N nodes, the time complexity per step is $O(N^2),$ and the algorithm usually takes a few steps to converge, while that of \cite{Bai17} is $O(N^3).$ However, we note that by using the Infection-immunization  algorithm \cite{BuloPB11} we can achieve even faster convergence as its per-step complexity would be linear in the number of nodes.

\section{Conclusion}
In this paper, we addressed a multi-feature fusion problem in CBIR. We developed a novel and computationally efficient CBIR method based on a constrained-clustering concept. In particular, we showed an efficient way of estimating a positive impact weight of features in a query-specific manner. Thus it can be readily used for feature combination. Furthermore, the proposed scheme is fully unsupervised, and can easily be able to detect false-positive NNs to the query, through the diffused similarity of the NNs. To demonstrate the validity of our method, we performed extensive experiments on benchmark datasets. Besides the improvements achieved on the state-of-the-art results, our method shows its effectiveness in quantifying the discriminative power of given features. Moreover, its effectiveness on feature-weighting can also be exploited in other computer vision problems, such as person re-identification, object detection, and image segmentation.

\section*{References}

\bibliography{egbib}
\end{document}